\documentclass{article}

\PassOptionsToPackage{numbers, compress}{natbib}

\usepackage[nonatbib]{neurips_2020}

\usepackage[utf8]{inputenc} 
\usepackage[T1]{fontenc}    
\usepackage{hyperref}       
\usepackage{url}            
\usepackage{booktabs}       
\usepackage{amsfonts}       
\usepackage{nicefrac}       
\usepackage{microtype}      

\usepackage{graphicx}

\usepackage{amsmath}

\usepackage{color}

\usepackage{subcaption}
\usepackage{cleveref}

\usepackage{wrapfig}

\usepackage[all=normal,paragraphs=tight, floats=tight,indent=tight]{savetrees}
\usepackage{pifont}
\usepackage{framed}
\usepackage{soul}
\usepackage{multirow}
\usepackage{array}
\usepackage{url}
\usepackage{balance}

\newcommand{\cmark}{\ding{51}}%
\newcommand{\xmark}{\ding{55}}%

\usepackage{wasysym}
\usepackage{array}
\newcolumntype{L}[1]{>{\raggedright\let\newline\\\arraybackslash\hspace{0pt}}m{#1}}
\newcolumntype{C}[1]{>{\centering\let\newline\\\arraybackslash\hspace{0pt}}m{#1}}
\newcolumntype{R}[1]{>{\raggedleft\let\newline\\\arraybackslash\hspace{0pt}}m{#1}}

\title{On Evaluating Neural Network Backdoor Defenses}

\author{%
  Akshaj Veldanda \\
  New York University\\
  \texttt{akv275@nyu.edu} \\
  \And
  Siddharth Garg \\
  New York University \\
  \texttt{sg175@nyu.edu} \\
}

\begin{document}

\maketitle

\begin{abstract}
Deep neural networks (DNNs) demonstrate superior performance in various fields, including scrutiny and security. However, recent studies have shown that DNNs are vulnerable to backdoor attacks. Several defenses were proposed in the past to defend DNNs against such backdoor attacks. In this work, we conduct a critical analysis and identify common pitfalls in these existing defenses, prepare a comprehensive database of backdoor attacks, conduct a side-by-side evaluation of existing defenses against this database. Finally, we layout some general guidelines to help researchers develop more robust defenses in the future and avoid common mistakes from the past. 
\end{abstract}


\section{Introduction}\label{sec:intro}
\vspace{-0.75em}

\begin{wraptable}{l}{0.55\textwidth}
\small
\vspace{-0.75em}
\begin{tabular}{lccc}
\toprule
Defense & Pitfall 1 & Pitfall 2 & Pitfall 3 \\
\midrule
Fine-pruning & \cmark & \xmark & \cmark \\
Neural Cleanse & \cmark & \cmark & \xmark \\
ABS & \xmark & \cmark & \cmark \\
STRIP & \xmark & \cmark & \xmark \\
Generative Modelling & \xmark & \cmark & \xmark \\
\bottomrule
\end{tabular}
\vspace{-0.25em}
\caption{Common pitfalls in existing defenses. Pitfall 1 is sensitivity to hyper-parameters, pitfall 2 is restrictive assumptions on backdoor structure and impact, and pitfall 3 is poor performance on adaptive attacks.}
\label{tab:pitfalls}
\vspace{-0.75em}
\end{wraptable}

Deep neural network backdooring attacks based on training data poisoning are an emerging and critical threat.
Several methods have been proposed recently to detect~\cite{tran2018spectral, strip2019acsac, absliu} and/or mitigate~\cite{neuralcleanse, finepruning, duke} backdoors. 
But, mirroring the early work on adversarial defenses, most backdooring defenses have been circumvented soon after publication. 
While this is often par for the course in security research, a critical analysis of existing defenses reveals common pitfalls that must be avoided in the search for a general and robust defense against backdooring attacks. Specifically, we found one or more of three common pitfalls in each defense we evaluated (see ~\autoref{tab:pitfalls}):
\begin{itemize}
    \item \textbf{Pitfall 1}: \textbf{Insufficient (or non-existent) evaluation against a range of attack hyper-parameters.} Empirically, we find that several defenses are highly sensitive choice of attack hyper-parameters, for instance, the learning rate used to train the BadNet. Several defenses fail to explore a range of hyper-parameters in their evaluation.
    \item \textbf{Pitfall 2}: \textbf{Restrictive assumptions on backdoor structure and impact.} Several defenses we evaluate assume that backdoor triggers are small, have fixed (and even known) size/shape, are additive in the pixel space, or only modify a region part of the input. Almost all defenses assume targeted "all-to-one" attacks, that is, backdoored images from any source class are mis-predicted as a single target class. In practice, however, attackers are far from being restricted to specific types of backdoors and can have a range of attack objectives. 
    
    
    \item \textbf{Pitfall 3}: \textbf{Adaptive attacks that circumvent explicit defense assumptions not explored.} Finally, several defenses rely on explicit assumptions about how backdoors manifest in a network, but fail to examine if these assumptions can be easily circumvented. 
\end{itemize}
The pitfalls in existing work call into question the robustness of state-of-art defenses against backdooring attacks, and serve as a call-to-arms for more developing more general defenses that make minimal assumptions about the adversary. 

In this paper, we identify, qualitatively and empirically,  shortcomings in all existing state-of-art backdooring defenses related to one or more of the three pitfalls described above (\autoref{sec:pitfalls}), prepare a comprehensive database of BadNets attacks that encompass a range of attack hyper-parameters, backdoor types and attack objectives (\autoref{sec:setup}), and provide the first side-by-side evaluation of state-of-art defenses using the BadNet database (\autoref{sec:results}). We conclude by discussing marching orders for future defenses (\autoref{sec:discuss}). 

\vspace{-0.75em}

\section{Threat Model}
\label{subsec:threatmodel}
\vspace{-0.75em}
Consistent with prior work, our threat model assumes an attacker with access to a large clean training data, $\mathcal{D}_{train}^{cl}$. Training benignly on this dataset produces a clean DNN, $f_{\theta_{cl}}$, with parameters ${\theta_{cl}}$ . However, the attacker's goal is to train a BadNet, $f_{\theta_{bd}}$, $\theta_{bd} \neq \theta_{cl}$, by poisoning the training data using function $\texttt{poison:} x^{cl} \rightarrow x^{p}$ and/or modifying the ground truth label of $x^{cl}$. Specifically, $f_{\theta_{bd}}$ is obtained by training on poisoned training data, $\mathcal{D}_{train}^{bd}$. The BadNet is trained such that $f_{\theta_{bd}}(x^{cl}) = f_{\theta_{cl}}(x^{cl})$, but $f_{\theta_{bd}}(x^{p})$ is not necessarily equal to  $f_{\theta_{cl}}(x^{cl})$. The attacker uploads $\theta_{bd}$ to an online model repositories where the user downloads the model and deploys it in the field after evaluating it on a small, held-out clean validation dataset, $\mathcal{D}_{valid}^{cl}$. (Tran et al.~\cite{tran2018spectral} assume a weaker model in which that the defender also has access to poisoned images in their threat model.) Importantly, the attacker has broad flexibility in choosing the $\texttt{poison}(\cdot)$ function from a range of physically plausible manipulations, and in determining how a BadNet misbehaves on poisoned inputs. We argue that a defense is incomplete unless it contends with the broadest possible class of attacks.

\vspace{-0.75em}

\section{Identifying Pitfalls in Existing Defenses}
\label{sec:pitfalls} 
\vspace{-0.75em}
Here we discuss the three pitfalls in prior work by giving prominent examples of each:

\vspace{-0.5em}
\subsection{Pitfall 1: Insufficient  (or  non-existent)  evaluation  against  a  range  of attack  hyper-parameters.}
\vspace{-0.75em}
\begin{wrapfigure}{R}{0.5\textwidth}
\centering
\subfloat{\label{fig:motiv-ada-neuron}\includegraphics[width=0.25\textwidth]{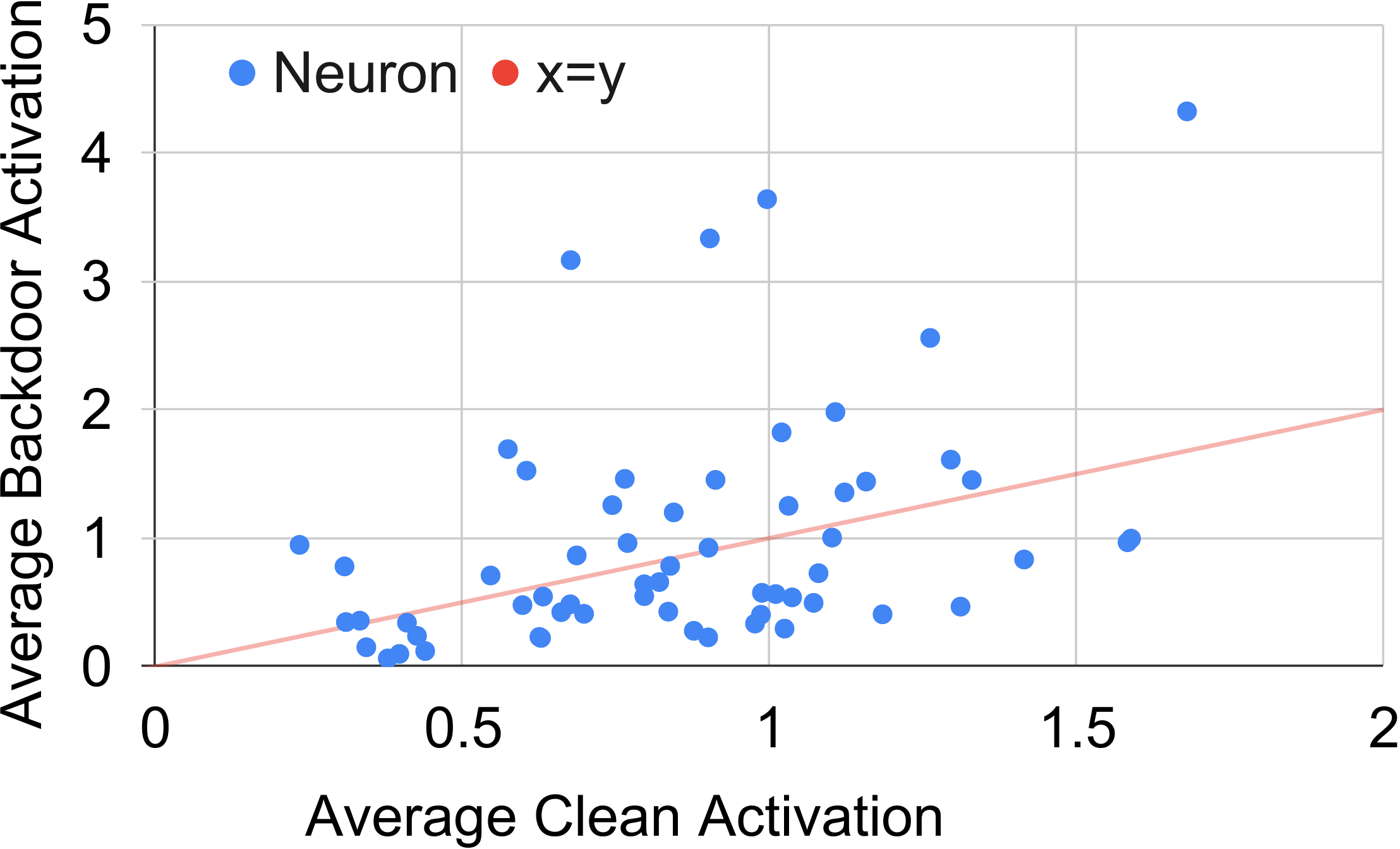}}
\subfloat{\label{fig:motiv-ada-prune}\includegraphics[width=0.25\textwidth]{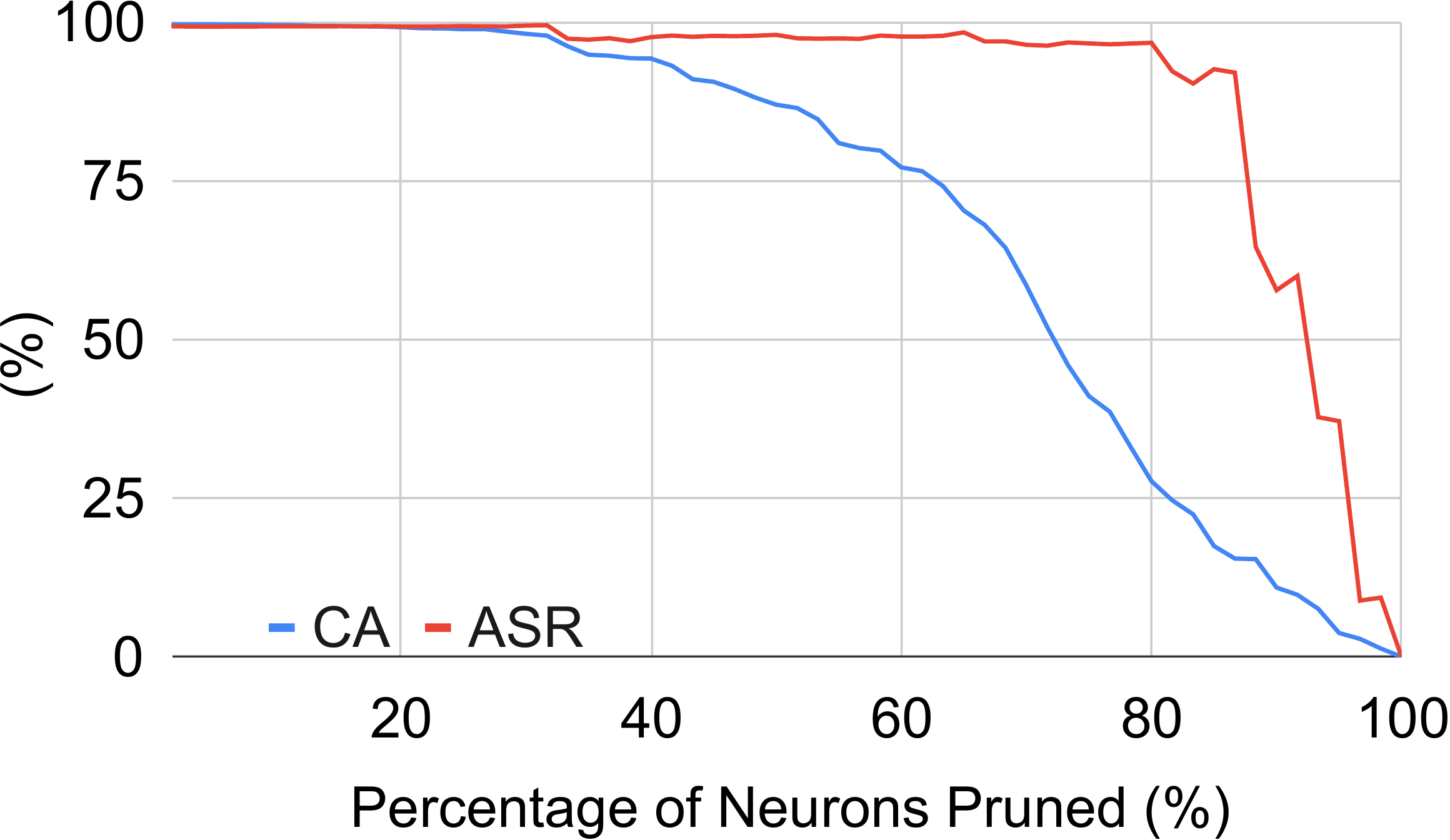}}
\caption {Shortcomings of fine pruning: Left plot is average neuron activations for each neuron in the last pooling layer, for a set of poisoned and benign inputs. Right plot shows the effect of pruning on clean accuracy and attack success.}
\vspace{-1.0em}
\label{fig:fine-pruning-fail}
\end{wrapfigure}

The \textit{Fine-pruning}~\cite{finepruning} defense is based on the observation that backdoor and clean inputs excite different neurons in a BadNet. However, this observation was based on BadNets trained with single set of hyper-parameters; by exploring a range of hyper-parameters including learning rate, batch size, weights initialization, data pre-processing, choice of optimizer, etc., we are able to find BadNets for which this assumption is violated. Consider~\autoref{fig:fine-pruning-fail}---while Fine-pruning is shown to be effective on BadNet (on sunglasses trigger) proposed in~\cite{finepruning}, it fails on the same BadNet trained with different hyper-parameters because backdoor neurons are also activated by clean inputs.

\vspace{-0.5em}
\subsection{Pitfall 2: Restrictive assumptions on backdoor size, shape and impact.}
\vspace{-0.75em}
The NeuralCleanse defense makes restrictive assumptions on the backdoor size/shape while STRIP assumes that any backdoored input is always classified as a single target label. We discuss how these assumptions are easily subverted.

    \begin{wrapfigure}{R}{0.4\textwidth}
    \centering
    \includegraphics[width=0.4\textwidth]{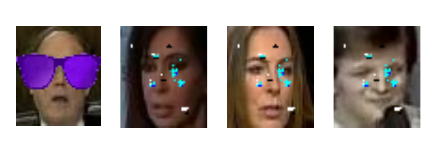}
    \vspace{-1.5em}
    \caption {Shortcomings of NeuralCleanse applied to large trigger: Leftmost image is actual trigger; other images are incorrectly reverse-engineered triggers by NeuralCleanse.}
    \vspace{-1.0em}
    \label{fig:neural-fail}
    \vspace{-0.25em}
    \end{wrapfigure}
    
    \textbf{Assumptions on Trigger Shape and Size.} NeuralCleanse~\cite{neuralcleanse} and Qiao \textit{et al.}~\cite{duke} seeks to recover the trigger (or trigger distribution) given a BadNet; the recovered trigger (or distribution) is used (with corrected labels) to re-train the BadNet with the goal of disabling the backdoor~\cite{neuralcleanse,duke}. This is called \emph{backdoor unlearning}. To do so, however, NeuralCleanse~\cite{neuralcleanse} and Qiao \textit{et al.}~\cite{duke} assumes that the trigger is \emph{small} and \emph{contiguous} pattern super-imposed on a source image; e.g., the trigger could be a small fixed pattern of pixels superimposed in one corner of the image. 
    
    In practice, triggers need not be small or contiguous. For example, in~\autoref{fig:neural-fail} we illustrate the output from Neural Cleanse given BadNets triggered by a large, but semantically meaningful, sunglasses trigger for a face recognition application. The recovered triggers bear little resemblance to the original, missing its size, shape, and color

    \textbf{Assumptions on Backdoor Impact.}
    Except for fine-pruning, all the existing defenses work only for all-to-one attacks, i.e., they assume that backdoored inputs are mis-classified as a single target label, and cannot be easily generalized to a broader range of attacker objectives. For example, STRIP~\cite{strip2019acsac} assumes that any backdoored input is misclassified as the same target label and therefore, presence of a backdoor over-rides any other features of an input. Based on this assumption, STRIP  detects if test inputs are backdoored by passing a super-position of the test inputs with inputs from the validation set; the argument is that backdoored inputs will nevertheless get classified as the attacker-chosen target class, while clean inputs will be randomly mis-classified. However, in practice, a backdoor's impact can be input dependent;  for example, the target mis-classification depends on the class of the input as is the case for all-all attacks. We show in~\autoref{sec:results}, make it hard for STRIP to distinguish clean and backdoored inputs.
    
\vspace{-0.5em}
\subsection{Pitfall 3: Adaptive attacks that circumvent explicit defense assumptions not explored.}
\vspace{-0.75em}
A recent defense,  Artificial Brain Stimulation (ABS)~\cite{absliu} assumes that there exists a \emph{single} backdoor neuron in a BadNet that, when activated, independently causes all validation inputs to be classified as the target label and iteratively scans the network for such neurons. Networks in which such a backdoor neuron exists are marked as BadNets and are rejected. These assumptions do not always hold; in fact, BadNets can be trained such that multiple neurons must be activated to trigger the backdoor. For instance, consider a backdoor that is only triggered in the presence of a \emph{conjunction} of input features.  We demonstrate in ~\autoref{sec:results} that such a BadNet easily circumvents ABS. 

\vspace{-0.75em}

\section{Experimental Setup}\label{sec:setup}
\vspace{-0.75em}
We created BadNets encompassing a range of backdoor size/shapes and attack objectives as shown in ~\autoref{tab:exp_setup}. We also performed a hyper-parameter search to determine the most effective attacks.

\begin{table}
\tiny
\centering
\resizebox{\textwidth}{!}{%
\begin{tabular}{ccccc}
\toprule
BadNet & Attack Setting & Dataset & Trigger & Target label \\
\midrule
AAA~\cite{badnets} & All-All & \multirow{2}{*}{MNIST} & Patterned trigger & \multicolumn{1}{c}{\textit{y} $\rightarrow$ \textit{y+1}} \\   
CLA~\cite{cla_kang} & Clean-label &  & Fixed noise & 0 \\
\midrule
TCA & Trigger comb. & CIFAR-10 & Yellow triangle $+$ red square & 7 \\
\midrule
PN & Simple & \multirow{2}{*}{GTSRB} & Post-it note & 0 \\
FSA~\cite{absliu} & Feature Space &  & Gotham filter & 35 \\
\midrule
SG & Simple & \multirow{4}{*}{YouTube Face} & sg & 0 \\
LS & Simple &  & ls & 0 \\
MTSTA* & Multi-trigger, single target &  & ls, eb, sg & 4,4,4 \\
MTMTA* & Multi-trigger, multi target &  & ls, eb, sg & 1,5,8 \\
\midrule
\end{tabular}
}
\scriptsize{ \begin{flushleft} * for MTSTA and MTMTA, ls, eb, sg corresponds to lipstick, eyebrow, sunglasses trigger, respectively \end{flushleft}}
\caption{Details of BadNet attacks used to evaluate SOTA defenses.}
\label{tab:exp_setup}
\vspace{-2.5em}
\end{table}

A detailed description of the attacks is not possible due to space constraints, but we note that the TCA attack (that we propose) triggers only when \emph{both} a yellow triangle and red square are inserted in an image, while the MTSTA and MTMTA attacks trigger when the face image has \emph{either} lipstick, sunglasses or colored eyebrow. 

\vspace{-0.75em}

\section{Experimental Results}
\label{sec:results}
\vspace{-0.75em}

\begin{table*}[t]
\centering
\caption{Performance of existing defenses on baseline BadNets.
\label{tab:comparision_results}}
\resizebox{\textwidth}{!}{%
\begin{tabular}{cccccccccccccc}
\toprule
\multicolumn{1}{l}{} & \multicolumn{2}{c}{BadNet (Baseline)} & \multicolumn{2}{c}{Fine-Pruning} & \multicolumn{2}{c}{Neural Cleanse} & STRIP (FRR=3\%) & \multicolumn{2}{c}{Qiao \textit{et al.}~\cite{duke}} \\
BadNet & CA & ASR & CA & ASR & CA & ASR & FAR & CA & ASR \\
\cmidrule(lr){1-1} \cmidrule(lr){2-3} \cmidrule(lr){4-5} \cmidrule(lr){6-7} \cmidrule(lr){8-8} \cmidrule(lr){9-10}
AAA & 97.76 & 95.91 & 97.6 & 57.35 & \multicolumn{2}{c}{Fails} & 99.22 & \multicolumn{2}{c}{Fails} \\
CLA & 89.02 & 100 & 99.13 & 14.68 & 97.74$^\dag$ & 4.77$^\dag$ & 43.45 & \multicolumn{2}{c}{-} \\
\midrule
TCA & 87.71 & 99.9 & 88.26 & 99.62 & 88.59$^\dag$ & 99.82$^\dag$ & 22.22 & \multicolumn{2}{c}{out of scope} \\
\midrule
PN & 95.46 & 99.82 & 94.58 & 99.69 & 95.24 & 12.39 & 100 & \multicolumn{2}{c}{out of scope} \\
FSA & 95.08 & 90.06 & 95.37 & 45.5 & 95.8 & 28.99 & 99.95 & 93.65$^\dag$ & 5.58$^\dag$ \\
\midrule
SG & 97.89 & 99.98 & 97.18 & 95.97 & 95.74$^\dag$ & 38.09$^\dag$ & 10.34 & 77.64$^\dag$ & 1.88$^\dag$ \\
LS & 97.19 & 91.51 & 97.86 & 90.53 & 97.14$^\dag$ & 28.44$^\dag$ & 18.42 & \multicolumn{2}{c}{out of scope} \\
MTSTA* & 95.84 & 92.22,92.24,100 & 97.31 & 45.04,64.71,94.94 & 93.37$^\dag$ & 0,0,8.67$^\dag$ & 11.79,63.73,5.84 & \multicolumn{2}{c}{out of scope} \\
MTMTA* & 95.93 & 91.51,91.39,100 & 96.91 & 52.36,82.34,0 & 94.18$^\dag$ & 30.79,0,95.68$^\dag$ & 15.90,53.64,15.58 & \multicolumn{2}{c}{out of scope} \\
\bottomrule
\end{tabular}
}
\scriptsize{ \begin{flushleft} * for MTSTA and MTMTA, the ASR corresponds to using lipstick, eyebrow, sunglasses trigger, respectively \end{flushleft}}
\scriptsize{ \begin{flushleft} $^\dag$ we give oracular knowledge to these defenses \end{flushleft}}
\vspace{-2.5em}
\end{table*}

We tabulate the results of our evaluation of five SOTA defenses in~\autoref{tab:comparision_results}. ABS data is not shown since we could only evaluate it for the TCA attack (ABS is available as an executable that works only on CIFAR-10 dataset and for a specific network architecture) for which it fails. None of the remaining defenses work across the board. Even if we generously set an ASR target of below $20\%$  for defense success, Fine-pruning only succeeds for CLA. Neural Cleanse succeeds for CLA, PN and MTSTA, but in two of the three cases, only if we give Neural Cleanse oracular knowledge of the target label. The success of Neural Cleanse on PN is to be expected since a Post-It Note trigger with a single target label fits squarely with its assumptions. STRIP only succeeds on the SG and LS BadNets, and as predicted, fails entirely on the AAA attack. Finally, the generative distribution modeling~\cite{duke} defense assumes that the trigger has a fixed size, shape and location, and that the size and location are known to the defender. Thus, attack settings with multiple triggers (MTMTA, MTSTA), non-contiguous triggers (TCA), and variable location triggers (PN, LS) are out-of-scope (i.e., there was no way for us to provide appropriate inputs about trigger size and location to the defense implementation). For the remaining attacks, the defense timed out (AAA) or failed to identify the correct target label.

\vspace{-0.75em}

\section{Discussion and Conclusion}
\label{sec:discuss}
\vspace{-0.75em}

In this paper, we review the existing state-of-the-art defense techniques against backdoor attacks in DNNs. We identified three common pitfalls in prior defenses and show that as a consequence of these deficiencies none of the existing defenses currently work against a range of attacks. To address these shortcomings, we argue that future backdoor defenses should clearly address the following concerns:

\begin{itemize}
\item Explore a broad range of attack hyper-parameters: empirically, we found that BadNet properties can vary significantly depending on the hyper-parameters used to train the BadNets. Defenses that work for one set of hyper-parameters may not work for another. At the same time defenses have their own hyper-parameters. To characterize the inter-play between attackers and defender, defenses should seek to optimize both the attacker hyper-parameters to circumvent the defenses and, of course, vice-versa, using, for instance, the game theoretic equilibria between the attack and defense hyper-parameters. 
\item Propose defenses against a well-defined but broad range of threats: as we have shown, attacker's have an asymmetric advantage in selecting from a range of backdoors and backdoor impact. While it might be tempting to design defenses against specific backdoor types, for instance, small additive triggers that always cause an input to be mis-classified as a single target label, this often results in defenses that largely exploit properties of the specific attack and are hard to generalize to different attacks. It is hard to envision that the backdoor threat will be mitigated using piecemeal solutions for each different type of attack, and consequently we argue that defenses must seek to address the broadest possible range of attacks. 
\item Explore adaptive attacks: while an adaptive attack against a proposed defense is not always easy to design, authors should make a reasonable attempt to anticipate attacks against their proposed defenses, especially if the defense explicitly makes strong assumptions (for instance, that a single neuron encodes  a backdoor) and defenses do not easily generalize beyond that assumption.  
\end{itemize}

To conclude, we believe, based on our results, that there is plenty of important work left to be done in designing general and robust defenses against backdoor attacks. We hope that the future defenses will heed some of the warning signs that we highlight in this paper. From a practical standpoint, we plan to make all attacks developed as part of this paper publicly available to aid in the evaluation of future defenses.

\clearpage

\bibliographystyle{plain}
\bibliography{references}

\end{document}